\documentclass[times,twocolumn,final,authoryear]{article}
\usepackage{simpleConference}
\usepackage{times}
\usepackage{graphicx}
\usepackage{amssymb}
\usepackage{url,hyperref}

\usepackage{natbib}
\usepackage[ngerman]{babel}
\usepackage{subcaption}
\usepackage{graphicx}
\usepackage{multirow}
\usepackage{array}
\usepackage{enumitem}
\setitemize{noitemsep,topsep=0.5pt,parsep=0.5pt,partopsep=0.5pt}
\newcolumntype{P}[1]{>{\centering\arraybackslash}p{#1}}
\usepackage{booktabs}

\usepackage{authblk}
\author[1]{Chengwei Wei}
\author[2]{Bin Wang}
\author[1]{C.-C. Jay Kuo}
\affil[1]{University of Southern California, Los Angeles, California, USA}
\affil[2]{National University of Singapore, Singapore}

{
    \makeatletter
    \renewcommand\AB@affilsepx{: \protect\Affilfont}
    \makeatother

    \makeatletter
    \renewcommand\AB@affilsepx{, \protect\Affilfont}
    \makeatother

    \affil[ ]{\texttt{chengwei@usc.edu}}
}

\begin{document}

\title{Task-Specific Dependency-based Word Embedding Methods}

% \author{Chengwei Wei\\
% \\
% University of Southern California \\
% Los Angeles, California, USA \\
% \today
% \\
% \\
% chengwei@usc.edu  \\
% }

\maketitle
\thispagestyle{empty}

\begin{abstract}
Two task-specific dependency-based word embedding methods are proposed
for text classification in this work. In contrast with universal word
embedding methods that work for generic tasks, we design task-specific
word embedding methods to offer better performance in a specific task.
Our methods follow the PPMI matrix factorization framework and derive
word contexts from the dependency parse tree.  The first one, called the
dependency-based word embedding (DWE), chooses keywords and neighbor
words of a target word in the dependency parse tree as contexts to
build the word-context matrix. The second method, named class-enhanced
dependency-based word embedding (CEDWE), learns from word-context as
well as word-class co-occurrence statistics. DWE and CEDWE are evaluated
on popular text classification datasets to demonstrate their
effectiveness.  It is shown by experimental results they outperform
several state-of-the-art word embedding methods. 
\end{abstract}

%%%%%% Introduction
\section{Introduction}\label{sec:introduction}

A word is represented by a real-valued vector through word embedding.
The technique finds applications in natural language processing (NLP)
tasks such as text classification, semantic search, parsing, and
machine translation \citep{zou2013bilingual,chen2014fast,liu2018task, de2016representation, wang2020sbert}.  Although contextualized word embedding
methods \citep{devlin2018bert, peters2018deep} offer state-of-the-art
performance, static word embedding methods \citep{levy2014neural, mikolov2013efficient} play a role because of their simplicity.  Static
word embedding methods can be categorized into count- and
prediction-based two types.  Positive point-wise mutual information
(PPMI) matrix factorization method \citep{levy2014neural} is a
count-based model, while the word2vec method \citep{mikolov2013efficient}
is a prediction-based model. GloVe \citep{pennington2014glove} is a
hybrid model consisting of both. Although these two models have
different model structures, both learn word embedding from the
co-occurrence information of words and their contexts. 

\begin{figure}[t]
%\centering
\includegraphics[width=0.9\columnwidth]{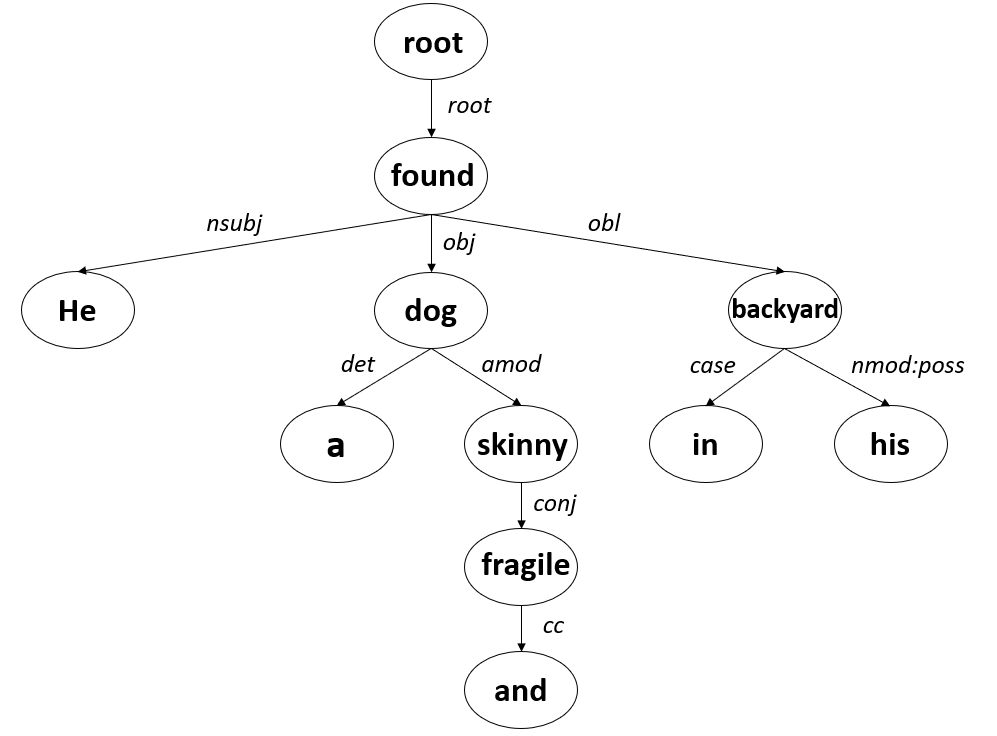} 
\caption{The dependency parse tree of an exemplary sentence - ``He found a skinny
and fragile dog in his backyard".}\label{fig:parsing_tree}
\end{figure}

Context selection is a key research topic in word representation
learning. Most word embedding methods adopt linear contexts.  For a
target word, its surrounding words are chosen as its contexts. The 
importance is ordered according to the distance.  The farther the
distance, the less the importance. An alternative is the
dependency-based context.  For each sentence, a syntactic dependency
parse tree can be generated by a dependency parser.  The neighbors of a
target word in the tree are chosen as its contexts. Dependency-based
contexts have been studied in count-based methods
\citep{pado2007dependency} and prediction-based methods
\citep{levy2014dependency}. As compared to linear contexts, dependency-based
contexts can find long-range contexts and exclude less informative
contexts. One example is shown in Fig. \ref{fig:parsing_tree}, where the target word
is `\emph{found}'.  Guided by the dependency parsing tree, its closely
related words (e.g.,  `\emph{he}', `\emph{dog}') can be easily
identified. In contrast, less related words (e.g.,  `\emph{skinny}',
`\emph{fragile}') are gathered by linear contexts. 

\renewcommand{\figurename}{Figure}
\renewcommand{\tablename}{Table}

%%%%%%%%%%%%%%%%%%%%%%%%%%%%%%%%%%%%%%%
%%%%%%%%%%%%%%%%%%%%%%%%%%%%%%%%%%%%%%%

Text classification, which assigns a class label to a sequence of texts,
is an application of the word embedding technique. One can compute the
text embedding from word embeddings and feed the information to the
classifier for class prediction.  One issue in most existing word
embedding models is that they only consider the contextual information
but do not take the task-specific information into account. The
task-specific information can be valuable for performance improvement.
It is worthwhile to mention that some word embedding methods were
proposed for text classification with improved performance by including
the topical information \citep{liu2015topical} or the syntactic
information \citep{komninos2016dependency}. 

Most recent work on dependency-based word embedding
\citep{levy2014dependency, komninos2016dependency} adopts word2vec's
skip-gram model.  It uses a target word to predict its contexts
constructed by dependency parsing. After parsing, we obtain triples from
the tree, each of which contains a head word, a dependent word, and the
dependency relationship between them.  For a target word, the
concatenations of its head or dependent words and their corresponding
dependency relation form dependency contexts (e.g., the contexts of
`\emph{found}' are \emph{he/nsubj}, \emph{dog/obj}, \emph{backyard/obl}
in Fig.  \ref{fig:parsing_tree}). Methods derived by this approach share
one common shortcoming.  Although they use a dependency parse tree to
construct contexts, all contexts are treated equally. Since the
dependency relation generated by a dependency parser captures the
syntactic information of a word in the sentence, more relevant contexts
can be extracted and exploited for more effective word embedding. 

In this work, two task-specific dependency-based word embedding methods
are proposed for text classification.  Our methods follow the PPMI
matrix factorization framework and derive word contexts based on the
dependency parse tree.  The first one, called the dependency-based word
embedding (DWE), chooses keywords and neighbor words of a target word in
the dependency parse tree as contexts to build the word-context matrix.
The second method, named class-enhanced dependency-based word embedding
(CEDWE), learns from word-context as well as word-class co-occurrence
statistics. DWE and CEDWE are evaluated on popular text classification
datasets to demonstrate their effectiveness.  It is shown by
experimental results they outperform several state-of-the-art word
embedding methods. 

There are three main contributions of this work as summarized below.
\begin{itemize}
\item We exploit the dependency relation in the dependency parse tree to
construct more effective contexts consisting of both keywords and
neighbor words. 
\item We propose a mechanism to merge word-context and word-class mutual
information into a single matrix for factorization so as to enhance text
classification accuracy. 
\item We conduct extensive experiments on large-scale text
classification datasets with the logistic regression and the XGBoost
classifiers to evaluate the effectiveness of the proposed DWE and CEDWE
methods. 
\end{itemize}

%%%%%% Related work
\section{Related Work}\label{sec:review}

Most word embedding methods learn the word representation with the
distributional hypothesis \citep{firth1957synopsis,
harris1954distributional}. That is, words with similar contexts are
expected to have similar meanings. It is natural to take the context
information into account in word embedding learning.  Several word
embedding methods were developed by following this idea.  Matrix
factorization methods \citep{levy2014neural} represent word contexts
using global corpus statistics. They construct a word-context
co-occurrence matrix and reduce its dimensionality.  Neural models such
as word2vec's skip-gram and CBOW \citep{mikolov2013efficient} learn word
embeddings by predicting contexts of the target word. GloVe combines the
two strategies and uses the gradient decent to reconstruct the global
co-occurrence matrix to learn word embeddings
\citep{pennington2014glove}.  Although these models appear different,
they share some similarities.  It was theoretically proved in
\citep{levy2014neural} that the learning process of ``skip-gram with
negative sampling (SGNS)'' actually factorizes a shifted PPMI matrix
implicitly. Further study in \citep{levy2015improving} offered a
connection between PPMI, skip-gram, and GloVe models. Experimental
results conducted on several intrinsic tasks indicate that none of these
models are significantly better than others. 

The syntactic information can be exploited in context construction to
learn better word embeddings. For example, research in
\citep{pado2007dependency} takes syntactic relations into account in
constructing the word-context co-occurrence matrix.  The syntactic
information was introduced to the skip-gram model in
\citep{levy2014dependency}. Furthermore, word embedding can be learned by
predicting the dependency-based context.  The second-order dependency
contexts were proposed in \citep{komninos2016dependency}.  In
\citep{li2018training}, weights were assigned to different dependencies
in the stochastic gradient descent process so that selected contexts are
not equally treated. More important contexts get higher weights. 

In this work, we focus on word embedding learning for the text
classification task. Word embedding methods have been tailored to text
classification for performance improvement. For example, the input text
can be classified into positive, negative, or neutral in sentiment
analysis. A neural model was proposed in \citep{tang2014learning} to
predict sentiment polarity in the word embedding training so that the
sentiment class information will have an impact on word vectors. It was
shown in \citep{komninos2016dependency} that the dependency-based word
embedding can improve the performance of the sentence classification
tasks because of the use of the syntactic information. Our work shares
some similarities with the task-oriented word embedding method proposed
in \citep{liu2018task}.  They modified skip-gram models by regularizing
the word class distribution to allow a clearer classification boundary
in the word embedding space. Here, we adopt the matrix factorization
method to learn embeddings from the word-context as well as the
word-class distribution statistics.

%%%%%% Methods
\section{Proposed DWE and CEDWE Methods}\label{sec:method}

We propose two new word embedding methods in this section. They are: DWE
(Dependency-based Word Embedding) and CEDWE (Class-Enhanced
Dependency-based Word Embedding). CEDWE is an enhanced version of DWE.
Both use the PPMI matrix factorization method as the basic word embedding
framework, which is briefly reviewed below.

%% add some formulas  
% The PPMI matrix factorization method counts the co-occurrence of
% word-context pairs in the training text and builds the PPMI matrix which
% is denoted by $X$.  Matrix $X$ is factorized with singular value
% decomposition (SVD) in form of
% \begin{equation}\label{eq:SVD}
% X= U \Sigma V,
% \end{equation}
% and the lower dimension matrix, $U \Sigma$, is adopted as the learned
% word embedding representation. 
Pointwise mutual information between a pair of word-context$(w,c)$ is defined as
\begin{equation}\label{eq:PMI0}
PMI(w,c)=log\frac{P(w,c)}{P(w)P(c)}
\end{equation}
where $P(w)$, $P(c)$ and $P(w,c)$ represent the probability of word $w$, context $c$ and joint probability of word $w$ and context $c$, respectively. 
The PMI can be estimated by
\begin{equation}\label{eq:PMI1}
PMI(w,c)=log\frac{N(w,c) \cdot |N|}{N(w) \cdot N(c)}
\end{equation}
where $N(w)$, $N(c)$ and $N(w,c)$ represent the number of $w$, $c$ and $(w,c)$ pair occur in corpus, respectively. $|N|$ is the total number of all possible $(w,c)$ pairs.

The PPMI matrix factorization method first counts the co-occurrence of
word-context pairs in the training text and estimate the PMI matrix.
Then the PPMI matrix $X$ is built by replacing all negative elements in the PMI matrix by 0.
\begin{equation}\label{eq:PPMI}
PPMI(w,c)=max(PMI(w,c), 0)
\end{equation}
Matrix $X$ is factorized with singular value
decomposition (SVD) in form of
\begin{equation}\label{eq:SVD}
X= U \Sigma V
\end{equation}
and the lower dimension matrix, $U \Sigma$, is adopted as the learned
word embedding representation. 
By following this framework, we study the use of dependency parsing to construct the word-context matrix. Then, we show how to enhance word embedding by exploiting word distributions in different classes.

\subsection{Contexts Selection in Dependency Parse Tree and DWE 
Method}\label{subsec:dependency}

Most previous work on dependency-based embeddings uses the concatenation
of the dependency relation and connected words as the context \citep{levy2014dependency, komninos2016dependency, li2018training}. This
choice increases the vocabulary size of contexts rapidly.  As a result,
the solution is not scalable with a large corpus size. Besides, it
demands larger memory space for matrix factorization. By focusing on
text classification, we can utilize dependency parsing to collect
related words for a target word and use them as contexts in constructing
the word-contexts co-occurrence matrix. The dependency relation can be
dropped. 

Our DWE method chooses two types of words in a dependency parse tree as
contexts. They are:
\begin{itemize}
\item Neighbor Words \\
The dependency parse tree for each sentence can be viewed as a graph. As
compared with linear contexts where contexts are surrounding words of
the target word, the dependency parse tree can offer informative
contexts from words at a farther distance.  We collect the words in the
$n$-hop neighborhood as contexts, where $n$ is a hyper-parameter. The
co-occurrence counts of words and contexts are weighted by their distance. 
\item Keywords \\
The main meaning of one sentence can generally be expressed by several
keywords in the sentence, such as subject, predicate, and object. Other
words have fewer impacts, such as function words. Keywords carry
important information of a sentence, although the diversity of semantic
meanings and syntactic structures constructed by keywords only is less
than those of the whole sentence. Generally speaking, constructing
contexts using keywords and paying more attention to them can provide
more informative and robust contexts. 
\end{itemize}

Each word has its own dependency relation with other words in a sentence
through dependency parsing.  Each relation represents a syntactic
relation of a dependent against its head word. It also
stands for the dependent word's syntactic function in the sentence.  

\begin{figure*}[t]
\centering
%%% Shrink
% \includegraphics[width=1\textwidth]{Figures/parsed sentence 1 final.png} 
\includegraphics[width=1\textwidth]{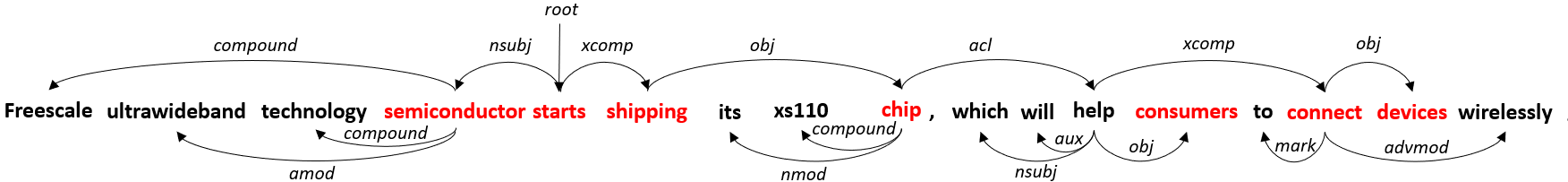}
\includegraphics[width=1\textwidth]{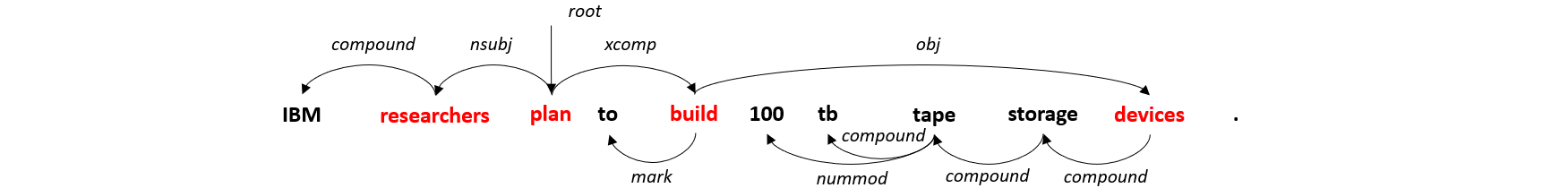} 
% \captionsetup{justification=centering}
\caption{
Three exemplary sentences represented by their dependency parse trees. The keywords spotted by our method are marked in red.
}
\label{fig:example}
\end{figure*}

%% Shrink
% Dependency relations have been classified by linguists into different
% categories based on their function, as shown in Table \ref{table1}
% \citep{de2014universal}.  We first exclude all stop words and punctuation
% in the dependency parse tree. Then, we locate dependent words whose
% dependency relations are in the core arguments (or the \emph{root}) and
% use them as keywords in the sentence. Examples are shown in Fig.
% \ref{fig:example}, where keywords found by our method are marked in red. The main parts are captured by the keywords, like \emph{"company"}, \emph{"hops"}, \emph{"woo"} and \emph{"makers"} in the first sentence, and \emph{"researchers"}, \emph{"plan"}, \emph{"build"} and \emph{"devices"} in the third sentence. Other words provide supplementary information to make the sentence complete, and they can be discarded without harming the main meaning of the sentence. For examples, in the first sentence, \emph{"music players"} defines what kind of \emph{"makers"} the company hops to woo, and in the third sentence, \emph{"100 tb tape storages"} give us more details about the \emph{"devices"}. Nevertheless, other words still contain information toward the sentence meaning. So they are used as contexts if they are in neighboring hops.
Dependency relations have been classified by linguists into different
categories based on their function, as shown in Table \ref{table1}
\citep{de2014universal}.  We first exclude all stop words and punctuations
in the dependency parse tree. Then, we locate dependent words whose
dependency relations are in the core arguments (or the \emph{root}) and
use them as keywords in the sentence. Examples are shown in Fig.
\ref{fig:example}, where keywords found by our method are marked in red. The main parts are captured by the keywords, like \emph{``semiconductor"}, \emph{``starts"}, \emph{``shipping"}, \emph{``chips"} and so forth in the first sentence, and \emph{``researchers"}, \emph{``plan"}, \emph{``build"} and \emph{``devices"} in the second sentence. Other words provide supplementary information to make the sentence complete, and they can be discarded without harming the main meaning of the sentence. For examples, in the second sentence, \emph{``100 tb tape storage"} give us more details about the \emph{``devices"}. Nevertheless, other words still contain information toward the sentence meaning. So they are used as contexts if they are in neighboring hops.
Besides the words in neighboring hops, keywords are always chosen as
part of contexts. All context words are weighted by the distance (in
terms of the number of hops) from the target word as defined in the
dependency parse tree. 

Counting co-occurrence with the keyword context provides a simple
context weighting scheme.  Note that some keywords may play a dual role.
That is, they may be neighbor words within $n$-hops as well. If this
occurs, we double the count of the word-keywords context so as to
increase its importance. 

%%%%%%%%%%%%%%%%%%%%%%%%%%%%%%%%%%%%%%%%%%%%%%%%
\begin{table}[t]
\centering
\includegraphics[width=0.9\columnwidth]{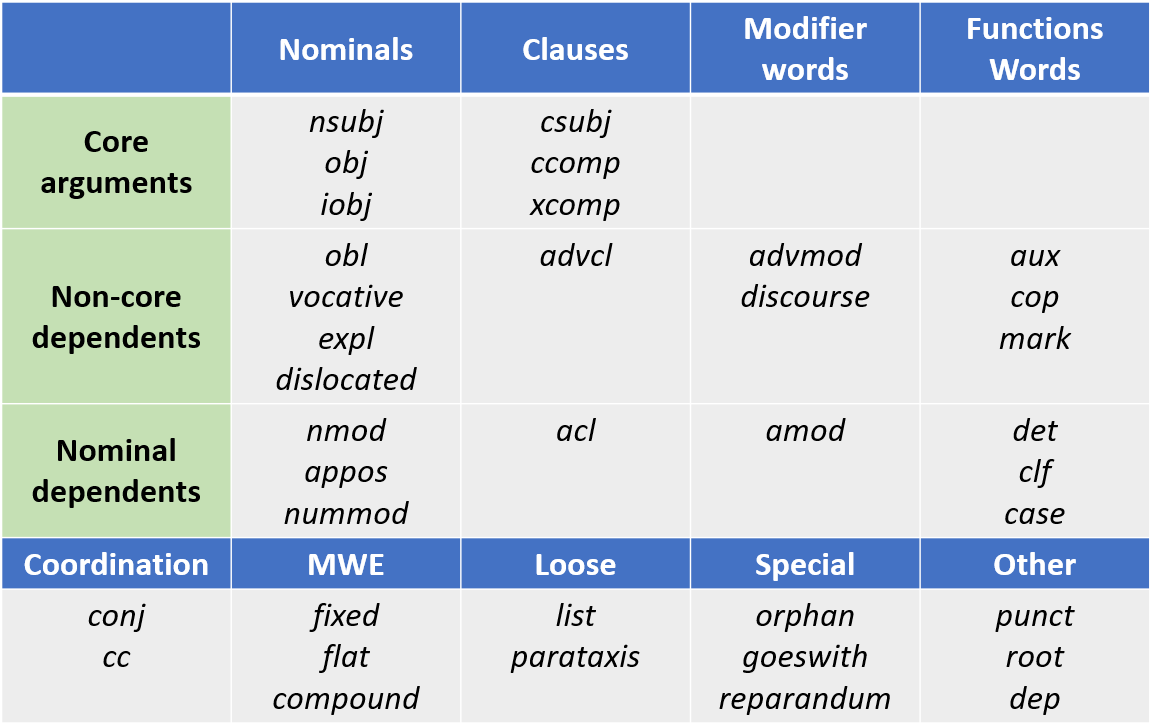}
\caption{Categories of universal syntactic relations. \citep{de2014universal}}
\label{table1}
\end{table}
%%%%%%%%%%%%%%%%%%%%%%%%%%%%%%%%%%%%%%%%%%%%%%%%

%%%%%%%%%%%%%%%%%%%%%%%%%%%%%%%%%%%%%%%%%%%%%%%%
%%%%%%%%%%%%%%%%%%%%%%%%%%%%%%%%%%%%%%%%%%%%%%%%

\subsection{Class-Enhanced Word-Context Matrix Construction and CEDWE 
Method}\label{subsec:class}

The word distribution in each class can vary in the text classification
task. That is, some words may have unique distributions in some classes.
By class-specific words, we mean those words that have special
distributions in specific classes. Class-specific words are one of the
prominent features in text classification.  To exploit class-specific
words, our idea is to build a word-class co-occurrence matrix and use
its row vectors as word features for classification. 

It is worthwhile to point out that, when we learn word embedding with
the word-contexts co-occurrence statistics, the class information is
incorporated in learned word embedding implicitly. This is because
contexts, which are essentially words, are distributed differently in
each class.  However, it is still possible to improve the classification
accuracy furthermore to use the word class distribution explicitly to
enhance the word embedding quality. This leads to the proposed
class-enhanced dependency-based word embedding (CEDWE) method. It is
detailed below.

To inject the class information into the word-context matrix, we modify
the raw word-context PPMI matrix, which is constructed by the whole
dataset, using the word distribution in each class. Generally speaking,
we compute the probabilities of words in each class and use them to
extend the row vectors of the PPMI matrix. Mathematically, ${\bf X}$ denotes
the raw word-context co-occurrence matrix. There are $n$ classes.  The
probability of word $i$ in each class is $p_i=[p_{i1}, p_{i2}, \cdots
p_{in}]$. For word $i$, we multiply its row vector, $X_{i,:}$, in the
PPMI matrix with its class probabilities and get an extended row vector
in form of
\begin{equation}\label{eq:embedding}
X'_{i,:} = [p_{i1} X_{i,:}, p_{i2} X_{i,:}, \cdots, p_{in} X_{i,:}],
\end{equation}
which is the row vector of the extended PPMI matrix ${\bf X'}$.  Then,
we apply SVD to the new matrix ${\bf X'}$ for dimensionality reduction
and get the learned word embeddings. 

After modifying the raw word-context PPMI matrix by the class
information, the new PPMI matrix contains both the word-context
information and the word-class information. The embedding of words that
appear frequently in the same classes are closer in the embedding space.
On the other hand, the embeddings for words that do not appear
frequently in the same classes are pulled away even they have similar
contexts. Then, the learned word embeddings are more suitable for
classification tasks. 

%%%%%% Experiments
\section{Experiments}\label{sec:experiments}

In this section, we conduct experiments to show the effectiveness of the
proposed DWE and CEDWE word embedding methods and benchmark them with
several other popular word embedding methods.  For DWE and CEDWE, texts
are parsed using the Stanza package \citep{qi2020stanza}.  Generally, the
classification performance of DWE and CEDWE increases as the hop number
becomes larger at the cost of higher complexity. Since the performance
improvement is limited after 3 hops, we search contexts inside the 3-hop
neighborhood to balance the performance/complexity trade-off. 

%%%%%%%%%%%%%%%%%%%%%%%%%%%%%%%%%%%%%%%%%%%%%%%%%%%%%%%%
\begin{table*}[htb]
\centering
\scalebox{0.9}{
\begin{tabular}{|l||l|l|l|l|l|l|l|} \hline
Word Embedding & AG\_NEWS & DBpedia & YahooAnswers & Yelp.P & Yelp.F & Amazon.P & Amazon.F\\ \hline
Word2vec & 89.08 & 96.63 & 68.21 & 88.77 & 52.95 & 84.30 & 46.87\\
GloVe & 89.64 & 96.85 & 67.78 & 86.79 & 51.10 & 82.38 & 44.76\\
EXT & 89.49 & 97.30 & 68.16 & 87.29 & 51.64 & 82.94 & 45.57\\
PPMI/LC & 89.87 & 97.33 & 69.22 & 91.19 & 55.12 & 87.13 & 48.55\\ \hline 
DWE (Ours) & \underbar{90.10} & \underbar{97.40} & \underbar{69.34} 
           & \underbar{91.30} & \underbar{55.23} & \underbar{87.47} & \underbar{48.58}\\
CEDWE (Ours) & \textbf{90.86} & \textbf{97.80} & \textbf{70.95} 
           & \textbf{92.94} & \textbf{57.63} & \textbf{89.68} & \textbf{52.48}\\ \hline
\end{tabular}
}
\caption{Test accuracy comparison of several word embedding methods 
with the Logistic Regression classifier, where the best and the second best
results are displayed in boldface and with underbar, respectively.}
\label{table:accuracy_logistic}
\end{table*}
%%%%%%%%%%%%%%%%%%%%%%%%%%%%%%%%%%%%%%%%%%%%%%%%%%%%%%%%

%%%%%%%%%%%%%%%%%%%%%%%%%%%%%%%%%%%%%%%%%%%%%%%%%%%%%%%%
\begin{table*}[t]
\centering
\scalebox{0.9}{
\begin{tabular}{|l||l|l|l|l|l|l|l|}
\hline
Word Embedding & AG\_NEWS & DBpedia & YahooAnswers & Yelp.P & Yelp.F & Amazon.P & Amazon.F\\
\hline
Word2vec & 89.71 & 96.60 & 68.37 & 88.23 & 51.77 & 84.46 & 46.25\\
GloVe & 90.63 & 96.87 & 68.57 & 86.67 & 49.89 & 82.77 & 44.54\\
EXT & 89.89 & 97.12 & 68.31 & 86.66 & 50.31 & 82.82 & 44.74\\
PPMI/LC & 90.65 & 97.36& 69.86 & 89.87 & 53.77 & 86.55 & 47.99\\ \hline 
DWE (Ours) & \underbar{90.87} & \underbar{97.45} & \underbar{69.95}
     & \underbar{90.11} & \underbar{54.22} & \underbar{86.78} & \underbar{48.06} \\
CEDWE (Ours) & \textbf{91.75} & \textbf{97.88} & \textbf{71.80} 
     & \textbf{92.19} & \textbf{57.03} & \textbf{89.28} & \textbf{51.85}\\ \hline
\end{tabular}
}
\caption{Test accuracy comparison of several word embedding methods 
with the XGBoost classifier, where the best and the second best
results are displayed in boldface and with underbar, respectively.}
\label{table:accuracy_xgboost}
\end{table*}
%%%%%%%%%%%%%%%%%%%%%%%%%%%%%%%%%%%%%%%%%%%%%%%%%%%%%%%%

\subsection{Datasets, Experiment Setup, and Benchmarks}

We adopt several large-scale text classification datasets from
\citep{zhang2015character} to train our word embedding methods and
conduct performance evaluation. If a random initialization is needed,
all results are obtained by averaging the results of 10 trials.
% shrink
\begin{itemize}
\item \textbf{AG\_NEWS}. AG\_NEWS is a 4-topic dataset extracted from
AG's corpus. Each topic has 30K training samples and 1.9K test samples. 
\item \textbf{DBpedia}. DBpedia is a project aiming to extract
structured content from the information in Wikipedia. The DBpedia text
classification dataset is constructed using 14 topics from DBpedia.
Each topic has 40K training samples and 5K test samples, where
each sample contains the title and abstract of an article. 
\item \textbf{YahooAnswers}. YahooAnswers is a 10-topic classification
dataset extracted from the Yahoo! Webscope program. Each topic has
140K training samples and 5K test samples. Each sample has a question and its answer. 
\item \textbf{YelpReviewPolarity \& YelpReviewFull}. Yelp review is a
sentiment classification dataset extracted from the 2015 Yelp Dataset
Challenge. It has full and polarity two versions.  YelpReviewFull has 5
classes ranging from stars 1 to 5, where each class has 130K training
samples and 10K test samples. In YelpReviewPolarity, stars 1 and 2
are treated as negative while stars 4 and 5 are viewed as positive. Each
class has 280K training samples and 19K test samples. 
\item \textbf{AmazonReviewPolarity \& AmazonReviewFull}. Amazon review
is also a sentiment classification dataset built upon Amazon customer
reviews and star rating. It has full and polarity two versions as well.
In AmazonReviewPolarity, each class has 600K training samples and 130K
test samples. In AmazonReviewFull, each class has 1.8M training samples
and 200K test samples. 
\end{itemize}

For each dataset, the most 50K frequent tokens with more than 10
occurrences are selected.  Punctuation and stops words are excluded.

We compare DWE and CEDWE with four other word embedding methods as
described below. 
\begin{itemize}
\item \textbf{Pre-train Word2vec} \citep{mikolov2013efficient}.  It is
trained on the Google News dataset using the SGNS model. 
\item \textbf{Pre-train GloVe} \citep{pennington2014glove}. We use the
GloVe.6b version trained on Wikipedia 2014 and Gigaword 5. 
\item \textbf{Extended Dependency Skip-gram (EXT)}
\citep{komninos2016dependency}. It is trained by the text classification
dataset using a dependency parse tree with the second-order dependency. 
\item \textbf{PPMI matrix with linear contexts (PPMI/LC)}. It is trained
by the text classification dataset with linear contexts for word-context
PPMI matrix construction followed by the SVD.  We set the window size to
10 so that its number of word-contexts is about the same as the number
of DWE. PPMI/LC is used to compare performance differences with linear
contexts and dependency-based contexts. 
\end{itemize}

We tested four word embedding dimensions (namely, 50, 100, 200, 300).  A
larger dimension often yields better performance in most datasets. We
will show embedding dimension impacts on performance later. For fair comparisons, we set the dimension of all word embedding
methods to 300.
% in test accuracy comparisons. 
We average the word embeddings in the text to get the text representation and compare the performance of two classifiers. They are the logistic regression classifier and the XGBoost. 

%%%%%%%%%%%%%%%%%%%%%%%%%%%%%%%%%%%%%%%%%%%%%%%%%%%%%%%%
\begin{table*}[t]
\centering
\scalebox{0.9}{
\begin{tabular}{ |l||P{2cm}|P{1.2cm}|P{1.2cm}|P{1.2cm}|P{1.2cm}|P{1.2cm}|P{1.2cm}|P{1.2cm}|  }\hline
& & AG & DBpedia & Y.A. & Yelp.P & Yelp.F & A.P & A.F\\
\hline
\multirow{2}{5em}{3-hop DWE w/o K} & Acc & 90.62 & 97.38 & 69.79 & 90.01 & 54.00 & 86.76 & 48.08 \\ 
 & (sample pairs) & (25.9M) & (130M) & (427M) & (254M) & (302M) & (930M) & (872M) \\ \hline
\multirow{2}{5em}{3-hop DWE} & Acc & 90.87 & 97.45 & 69.95 & 90.11 & 54.22 & 86.78 & 48.06 \\ 
 & (sample pairs) & (33.9M) & (141M) & (516M) & (293M) & (348M) & (1077M) & (1011M) \\ \hline
\multirow{2}{5em}{5-hop DWE w/o K} & Acc & 90.80 & 97.47 & 70.03 & 90.21 & 54.23 & 86.80 & 48.11 \\ 
 & (sample pairs) & (47.4M) & (200M) & (660M) & (372M) & (442M) & (1348M) & (1269M)\\ \hline
\end{tabular}
}
\caption{Classification accuracy results and the number of word-context
sample pairs (in the unit of million) for the dependency-based contexts,
where ``DWE w/o K" means the proposed DWE method without the use of the
keyword context.} \label{table:ablation}
\end{table*}
%%%%%%%%%%%%%%%%%%%%%%%%%%%%%%%%%%%%%%%%%%%%%%%%%%%%%%%%

%%%%%%%%%%%%%%%%%%%%%%%%%%%%%%%%%%%%%%%%%%%%%%%%%%%%%%%%
\begin{figure*}[ht]
\centering % <-- added
\begin{subfigure}{0.3\textwidth}
  \includegraphics[width=\linewidth]{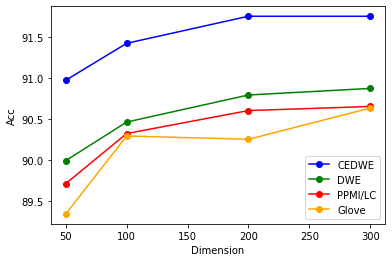}
  \caption{AG\_NEWS}
  \label{fig:d_ag}
\end{subfigure}\hfil % <-- added
\begin{subfigure}{0.3\textwidth}
  \includegraphics[width=\linewidth]{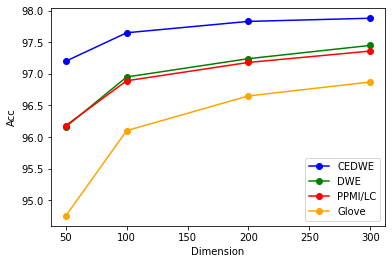}
  \caption{DBpedia}
  \label{fig:d_dbpedia}
\end{subfigure}\hfil % <-- added
\begin{subfigure}{0.3\textwidth}
  \includegraphics[width=\linewidth]{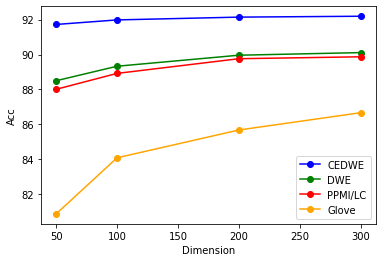}
  \caption{YelpReviewPolarity}
  \label{fig:d_yelpreviewpolarity}
\end{subfigure}\hfil % <-- added
\caption{The classification accuracy curves as a function of embedding dimensions
for three datasets: (a) AG\_NEWS, (b) DBpedia and (c) YelpReviewPolarity,
where the tested dimensions are set to 50, 100, 200 and 300.}\label{fig:dimension}
\end{figure*}
%%%%%%%%%%%%%%%%%%%%%%%%%%%%%%%%%%%%%%%%%%%%%%%%%%%%%%%%

%%%%%%%%%%%%%%%%%%%%%%%%%%%%%%%%%%%%%%%%%%%%%%%%%%%%%%%%
\begin{figure}[ht]
\centering 
\includegraphics[width=0.3\textwidth]{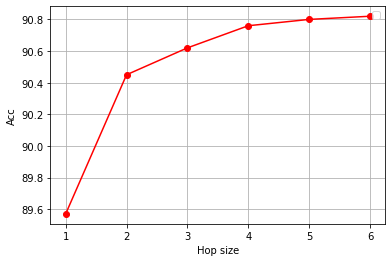}
\centering
\caption{The classification accuracy as a function of hop sizes for the
AG\_NEWS dataset, where the results are obtained by using only $n$-hop
neighbor words in the dependency parse tree as contexts (namely, the
keyword contexts are ignored), where $n=1, \cdots, 6$.}\label{fig:hop}
\end{figure}
%%%%%%%%%%%%%%%%%%%%%%%%%%%%%%%%%%%%%%%%%%%%%%%%%%%%%%%%

%%%%%%%%%%%%%%%%%%%%%%%%%%%%%%%%%%%%%%%%%%%%%%%%%%%%%%%%
\begin{figure}[t]
\centering % <-- added
\begin{subfigure}{0.24\textwidth}
\includegraphics[width=\linewidth]{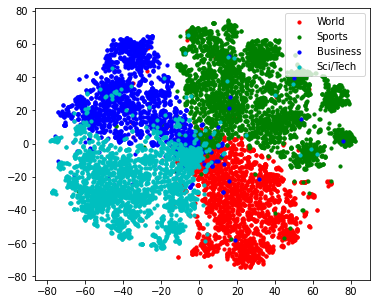}
\caption{AG\_NEWS: DWE}\label{fig:1}
\end{subfigure}\hfil % <-- added
\begin{subfigure}{0.24\textwidth}
\includegraphics[width=\linewidth]{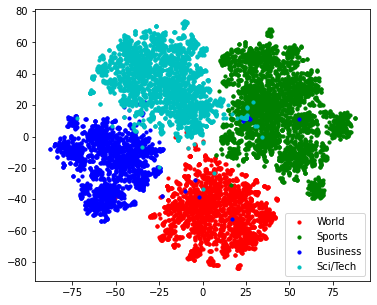}
\caption{AG\_NEWS: CEDWE}\label{fig:2}
\end{subfigure}\hfil % <-- added
\caption{Visualization of the embedding spaces of (a) DWE and (b) CEDWE 
for the AG\_NEWS dataset.}\label{fig:visual_ag}
\begin{subfigure}{0.235\textwidth}
\includegraphics[width=\linewidth]{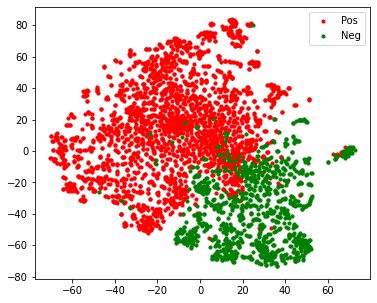}
\caption{YelpReviewPolarity: DWE}\label{fig:3}
\end{subfigure}\hfil % <-- added
\begin{subfigure}{0.245\textwidth}
\includegraphics[width=\linewidth]{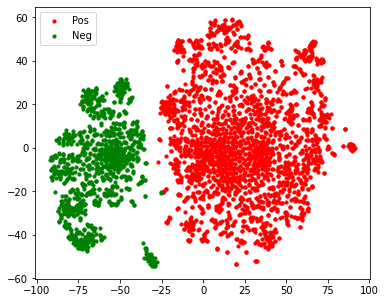}
\caption{YelpReviewPolarity: CEDWE}\label{fig:4}
\end{subfigure}\hfil % <-- added
\caption{Visualization of the embedding spaces of (a) DWE and (b) CEDWE 
for the YelpReviewPolarity dataset.}\label{fig:visual_yelp}
\end{figure}
%%%%%%%%%%%%%%%%%%%%%%%%%%%%%%%%%%%%%%%%%%%%%%%%%%%%%%%%

%%%%%%%%%%%%%%%%%%%%%%%%%%%%%%%%%%%%%%%%%%%%%%%%%%%%%%%%
% \begin{figure*}[t]
% \centering % <-- added
% \begin{subfigure}{0.24\textwidth}
%   \includegraphics[width=\linewidth]{Figures/YRP_DCK.png}
%   \caption{YelpReviewPolarity: DWE}
%   \label{fig:3}
% \end{subfigure}\hfil % <-- added
% \begin{subfigure}{0.24\textwidth}
%   \includegraphics[width=\linewidth]{Figures/YRP_DCK_CLASS.png}
%   \caption{YelpReviewPolarity: CEDWE}
%   \label{fig:4}
% \end{subfigure}\hfil % <-- added
% \caption{Visualization of the embedding spaces of (a) DWE and (b) CEDWE 
% for the YelpReviewPolarity dataset.}\label{fig:visual_yelp}
% \end{figure*}
%%%%%%%%%%%%%%%%%%%%%%%%%%%%%%%%%%%%%%%%%%%%%%%%%%%%%%%%

\subsection{Results and Analysis}

%{\bf Test Accuracy.} 
Experimental results with the logistic
regression classifier and the XGBoost classifier are shown in Tables
\ref{table:accuracy_logistic} and \ref{table:accuracy_xgboost},
respectively. Word2vec, GloVe, and EXT are trained on general
large-scale corpora which have more than billions of tokens.  In
contrast, PPMI/LC, DWE and CEDWE are directly trained on the text
classification dataset. The contexts from the corresponding
classification dataset makes them suitable for the following
classification task.  We see some performance gain of word embeddings
trained on the text classification dataset (i.e., the last three rows)
over pre-trained word embedding methods (i.e., the first three rows).
Such gains are especially obvious with the two proposed methods. DWE
outperforms PPMI/LC consistently. After incorporating the class
information in the word embedding mechanism, CEDWE achieves the best
performance.  The performance improvement of DWE and CEDWE is primarily
due to the design of a word embedding method to match its target task.
We see the benefit of re-training or fine-tuning of a word embedding
scheme in specific tasks. 

{\bf Effect of Embedding Dimension.} Generally, word embeddings of a
larger dimension have better classification performance. We show the
classification accuracy curves as a function of word embedding
dimensions with the XGBoost classifier in Fig.  \ref{fig:dimension}.
Furthermore, we see a significant performance gap between CEDWE and
other word embeddings when the dimension is lower. We can see our
proposed method also performs well when the dimension is low. It
indicates that the proposed CEDWE is a good choice when a lightweight
model is essential in an application scenario. 

{\bf Linear vs. Dependency-based Contexts.} We see from Tables
\ref{table:accuracy_logistic} and \ref{table:accuracy_xgboost} that
there is a clear performance gap between the PPMI/LC and DWE. As
compared with general word embeddings trained on large-scale corpora,
word embedding methods trained for specific tasks usually have much less
training texts. For such an environment, dependency-based contexts are
more informative than linear contexts. The use of a syntactic dependency
parser can make obtained contexts more robust and stable. 

{\bf Effect of Keyword Contexts.} It is observed in our experiments that
the classification performance increases as the neighbor-hop size in
dependency-based contexts (or the window size in linear contexts)
increases. This is because more word-context pairs are collected with a
larger hop number (or window size).  Nevertheless, the improvement is
diminishing when the hop size (or the window size) reaches a certain
level as illustrated in Fig. \ref{fig:hop}, where the classification
accuracy is plotted as a function of the hop size for the AG\_NEWS
dataset.

Interestingly, we can leverage keywords and use them as extra contexts
to allow a smaller hop size to reduce the computational complexity as
shown in Table \ref{table:ablation}.  We compare three ways to choose
word contexts based on the dependency parse tree: 
\begin{enumerate}
\item DWE with 3-hop neighbor contexts only;
\item DWE with 5-hop neighbor contexts only;
\item DWE with 3-hop neighbor contexts and keyword contexts;
\end{enumerate}
Since the default DWE has both neighbor and keyword contexts, we use the
notation ``DWE w/o K" (DWE without keyword contexts) to denote the first
two cases. The classification accuracy results reported in the table are
obtained using the XGBoost classifier. For AmazonReview dataset, the number of sample word-context pairs is already enough for hop size 3 and using more word-context pairs won't increase the performance. 
For AG\_NEWS, DBpedia, YahooAnswers, and YelpReviewFull datasets, we can see the effectiveness of using keywords as additional contexts. 
The performance of the second
and third cases is close to each other while the number of sample pairs
in the third case is significantly smaller than that in the second case. 

{\bf Effect of Explicit Class Information.} Some words have similar
contexts but appear in different classes in text classification. For
example, adjectives in different classes can modify the same object in
movie review datasets (e.g., ``a nice movie'', ``a funny movie'', ``a
disappointed movie'', ``a terrible movie'').  General word embedding
methods may have these adjectives closer since they have some similar
contexts. This is, however, undesirable for classification tasks.  The
proposed CEDWE method takes the word class information into account in
forming the word-context PPMI matrix to address this shortcoming.  In
the embedding space, the boundaries of class-specific words of different
classes becomes clearer and words that frequently appear in the same
class are pulled together. 

Task-specific words frequently appear in some specific classes so that
they have higher occurrence probabilities in the corresponding classes.
We use the chi-square test to select class-specific words and denote
them with the class that has the highest occurrence probability. Then,
t-SNE dimensionality reduction is used to visualize these task-specific
words in the embedding space.  The t-SNE plots of the embedding spaces
of DWE and CEDWE for AG\_NEWS and YelpReviewPolarity are shown in Figs.
\ref{fig:visual_ag} and \ref{fig:visual_yelp}, respectively.  As
compared with DWE, class-specific words in different classes are better
separated in CEDWE. This explains why CEDWE has better classification
performance than DWE. 

%%%%%% Conclusion
\section{Conclusion and Future Work}\label{sec:conclusion}

Two dependency-based word embedding methods, DWE and CEDWE, were
proposed in this work.  DWE uses keywords in sentences as extra contexts
to build the word-context matrix. It provides informative contexts in a
larger scope.  As a result, as compared with the scheme that only uses
neighbor words as contexts, it achieves comparable text classification
performance with less word-context sample pairs. To improve the text
classification performance furthermore, CEDWE incorporates the word
class distribution.  The t-SNE plot visualization tool was introduced to
explain the superior performance of CEDWE. 

As future extensions, it would be interesting to exploit more
well-defined weighting function on contexts based on the dependency
relation. It is also worthwhile to learn effective word embedding
methods for intrinsic and extrinsic tasks that go beyond text
classification. 

\bibliographystyle{plainnat}
\renewcommand\refname{Reference}
\bibliography{mybibfile}

\end{document}